\documentclass[runningheads]{llncs}

\usepackage{eccv}

\usepackage{eccvabbrv}

\usepackage{graphicx}
\usepackage{booktabs}
\usepackage{algorithm,algorithmic}

\usepackage[accsupp]{axessibility}  

\usepackage[pagebackref,breaklinks,colorlinks]{hyperref}

\usepackage{orcidlink}

\begin{document}

\title{Self-supervised co-salient object detection via feature correspondence at multiple scales} 

\titlerunning{Self-supervised CoSOD via multi-scale feature correspondences}

\author{Souradeep Chakraborty \and
Dimitris Samaras} 

\authorrunning{Chakraborty et al.}

\institute{Department of Computer Science, Stony Brook University, New York, USA
\email{\{souchakrabor,samaras\}@cs.stonybrook.edu}}

\maketitle

\begin{abstract}
Our paper introduces a novel two-stage self-supervised approach for detecting co-occurring salient objects (CoSOD) in image groups without requiring segmentation annotations. Unlike existing unsupervised methods that rely solely on patch-level information (\eg clustering patch descriptors) or on computation heavy off-the-shelf components for CoSOD, our lightweight model leverages feature correspondences at both patch and region levels, significantly improving prediction performance. In the first stage, we train a self-supervised network that detects co-salient regions by computing local patch-level feature  correspondences across images. We  obtain the segmentation predictions using confidence-based adaptive thresholding. In the next stage, we refine these intermediate segmentations by eliminating the detected regions (within each image) whose averaged feature representations are dissimilar to the foreground feature representation averaged across all the cross-attention maps (from the previous stage). Extensive experiments on three CoSOD benchmark datasets show that our self-supervised model outperforms the corresponding state-of-the-art models by a huge margin (\eg, on the CoCA dataset, our model has a 13.7\% F-measure gain over the  SOTA unsupervised CoSOD model). Notably, our self-supervised model also outperforms several recent fully supervised CoSOD models on the three test datasets (\eg, on the CoCA dataset,  our model has a 4.6\% F-measure gain over a recent supervised CoSOD model). \\

Project page: \textcolor{blue}{https://github.com/sourachakra/SCoSPARC}

\end{abstract}    
\vspace{-4mm}
\section{Introduction}
\label{sec:intro}
Co-salient object detection (CoSOD) identifies co-existing salient objects among a collection of images, leveraging shared semantic information across image regions within the group, resulting in more accurate localization compared to single-image salient object detection (SOD) models \cite{chen2020global,li2021uncertainty,liu2021samnet,piao2021mfnet,tang2021disentangled,yu2021structure,li2021salient}. Both tasks, CoSOD and SOD, encompass joint segmentation and detection activities, necessitating segmentation labels, which are resource-intensive to acquire due to their time-consuming nature, as evidenced in the existing literature \cite{yu2022democracy,fan2021re,fan2021group}.

\begin{figure}
\centering
\includegraphics[width = 9.2cm]{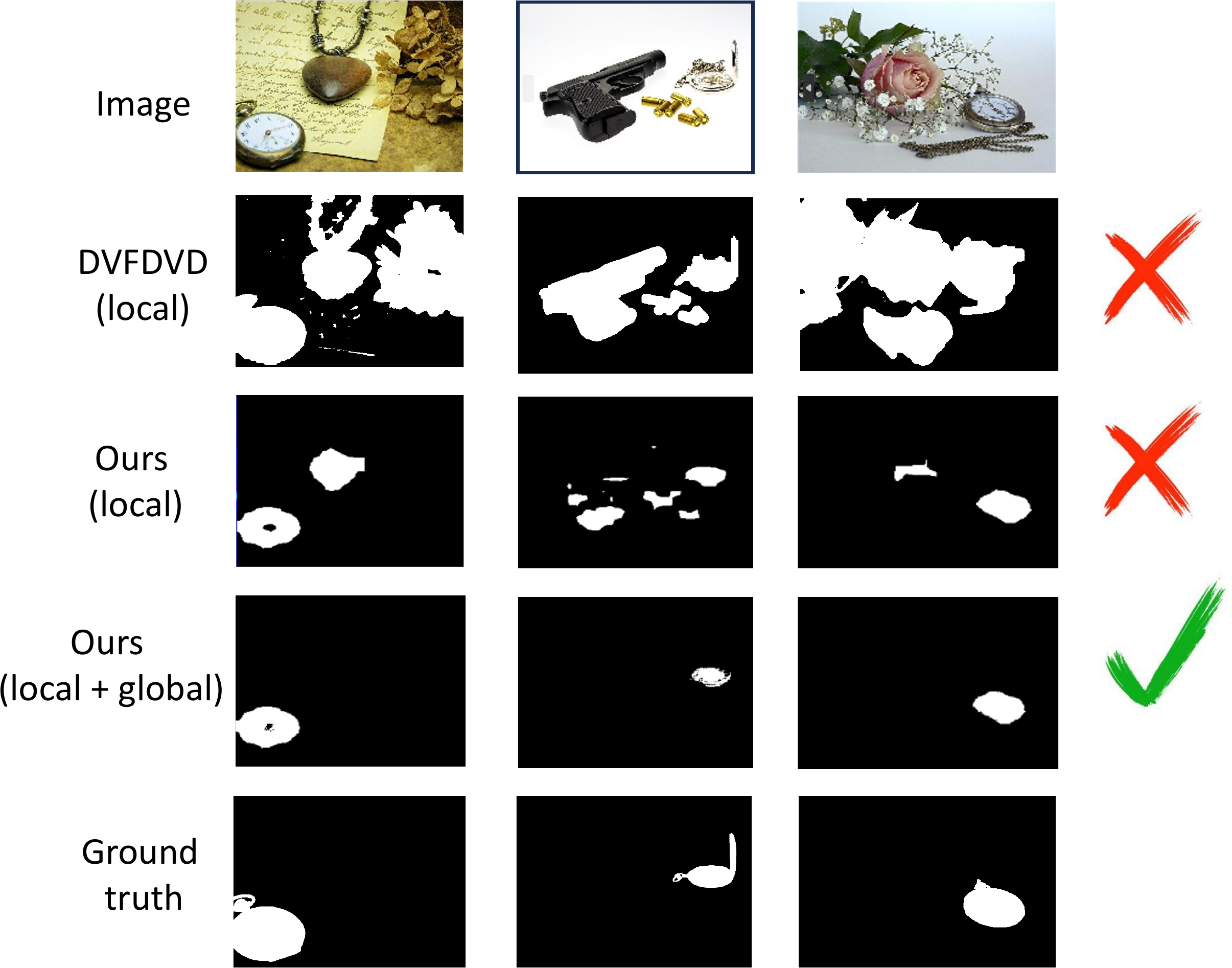}
\caption {Visualization of co-saliency detections on the \textit{pocket watch} image
group from the CoCA dataset \cite{zhang2020gradient}: Row 1: Original image, Row 2: predictions from the DVFDVD model \cite{amir2021deep} that only mines local patch-level correspondences, Row 3: our predictions with only local (patch) feature correspondence, Row 4: our predictions with both local (patch) and global (region) feature correspondence which produces the best results, Row 5: Ground truth.}
\label{fig:teaser}
\vspace{-2.5mm}
\end{figure}

The reliance on annotations poses a challenge for the existing fully supervised CoSOD models \cite{fan2021group,zhang2021deepacg,zhang2020gradient,fan2021re,yu2022democracy}. To alleviate this burden, certain approaches \cite{li2018unsupervised,hsu2018co,hsu2018unsupervised} have focused on unsupervised co-segmentation and co-saliency detection, having several potential real-world applications such as e-commerce, content-based image retrieval, satellite imaging, biomedical imaging, etc. However, these models demonstrate significantly poorer prediction performance compared to their fully supervised counterparts due to their inefficient utilization of unlabeled data at different scales (patch and region levels). For example, Amir et al.~\cite{amir2021deep} only mines local patch-level features such as clustering of ViT patch descriptors for co-segmentation. Also, for some models \cite{xiao2023zero, Chakraborty_2024_WACV} the performance improvement comes mostly from using heavy off-the-shelf components (such as SAM \cite{kirillov2023segment}, STEGO \cite{hamilton2022unsupervised}, DCFM \cite{yu2022democracy}, Stable Diffusion \cite{rombach2022high}, etc.) that make their models computation-heavy and hence unfit for real-time applications.

In this paper, we present a lightweight model that leverages feature correspondences at both patch and region levels to improve unsupervised CoSOD performance. We take advantage of the self-supervised features in visual transformers (ViTs) \cite{dosovitskiy2020image,caron2021emerging} and their self-attention maps \cite{caron2021emerging} to develop a simple yet effective self-supervised CoSOD model that uses both patch and region level feature correspondences, which we call SCoSPARC. 

As part of our self-supervised approach, we first train a network to compute cross-attention maps that highlight commonly occurring salient regions via local patch-level feature correspondences across images in the group. We show that these correspondences form strong signals for unsupervised CoSOD. We design this network to optimize two losses: 1) the Co-occurrence loss, which constraints  the foreground image regions to have similar feature representations and at  the same time, it forces the foreground and background feature representations within each image to be as dissimilar as possible, 2) the Saliency loss, in order to maximize the saliency of the detected regions. While previous approaches \cite{hsu2018unsupervised,hsu2018co}, have also leveraged similar losses for guiding their model training, we differ from these approaches in two main aspects: 1) we avoid the use of separate off-the-shelf saliency models for training, instead directly leverage the intermediate self-attention maps from our backbone encoder to construct our saliency signal, 2) we construct  the foreground and the background feature embeddings directly using  the feature descriptors from our backbone encoder (averaged via cross-attention maps) instead of training separate sub-networks to extract the mask embeddings (details in  Sec.~\ref{sec:sec31}). Thus, we   effectively leverage the feature encodings from our backbone encoder to construct both of our  co-occurrence and saliency signals during training. This helps us maintain our model's computational efficiency and facilitates fast inference times. Next, we introduce a prediction confidence-based adaptive thresholding method for thresholding the cross-attention maps in order to generate intermediate CoSOD segmentations. While prior works \cite{wang2022freematch,guo2022class} have used adaptive thresholding based on the class confidence in the context of semi-supervised learning, we use adaptive thresholds based on the  confidence of the co-saliency predictions, for our task of self-supervised CoSOD. Our model at this stage outperforms the SOTA unsupervised US-CoSOD model \cite{Chakraborty_2024_WACV} by a significant margin (details in  Sec.~\ref{sec:results}). For enforcing region-level feature correspondence, we next identify connected components (corresponding to the detected image regions) in the intermediate segmentation masks and eliminate regions whose feature representations are not similar to the average foreground feature representation obtained from the cross-attention maps in the previous step. Our experiments demonstrate a significant improvement in prediction performance compared to existing works. 

In Fig.~\ref{fig:teaser}, we show CoSOD predictions from three different methods: 1) the DVFDVD model \cite{amir2021deep} that only mines local patch-level feature  correspondences via clustering of patch descriptors, 2) our predictions with only local (patch-level) feature correspondence, 3) our predictions with both local (patch-level) and more global (region-level)  feature correspondences, which produces the best results. Fig.~\ref{fig:teaser} demonstrates the two main contributions of our work: 1) our local patch-level feature correspondence learning network produces better results compared to the existing models \eg DVFDVD~\cite{amir2021deep} which only clusters the local ViT patch descriptors, 2) using both local (patch-level) and global (region-level) feature correspondences for CoSOD helps improve the results.

\noindent We summarize our main contributions as follows:
\vspace{-1mm}
\begin{itemize}
  \item We propose a simple yet effective two-stage self-supervised approach for CoSOD that leverages feature correspondences (of self-supervised ViT features) at different scales in an image group.
    \vspace{1mm}
  \item We introduce a confidence-based adaptive thresholding approach for the cross-attention maps, outperforming the conventional fixed threshold of 0.5 commonly used in binary segmentation tasks.
   \vspace{1mm}
  \item We show that our method outperforms existing unsupervised CoSOD approaches on three benchmarks (\eg, on the CoCA dataset, our model has a 13.7\% F-measure gain over the SOTA unsupervised CoSOD model) while also outperforming several popular recent supervised CoSOD methods on these datasets. 

\end{itemize}

\section{Related Work}
\label{sec:formatting}

\textbf{Self-supervised learning:} Unlike supervised methods that necessitate human annotation, self-supervised learning involves training networks with automatically generated pseudo-labels that capture characteristics such as image contexts or handcrafted cues in order to accomplish a pretext task (\eg colorization, rotation prediction, etc.) using unlabeled data \cite{caron2021emerging,chen2020simple,gidaris2021obow,he2020momentum,noroozi2016unsupervised,tian2020makes}. For instance, DINO \cite{caron2021emerging} employs a student-teacher framework where the two networks observe different and randomly transformed input parts, and the student network learns to predict the mean-centered output of the teacher network. Studies based on the DINO ViT features have leveraged these features for tasks such as object discovery \cite{simeoni2021localizing,wang2022self}, semantic segmentation \cite{hamilton2022unsupervised,van2021unsupervised}, and category discovery \cite{vaze2022generalized}. In Masked Auto Encoder \cite{he2022masked}, patches of the input image are randomly masked, and the pretext task involves learning to reconstruct the missing pixels through auto-encoding. These studies have demonstrated that the representations derived from the self-attention maps of ViTs contain valuable localization information \cite{amir2021deep,caron2021emerging,zhou2021ibot}. Our work incorporates both the patch-level ViT feature descriptors and the self-attention maps from DINO to guide our self-supervised network training.

\textbf{Co-salient object detection:}
Graphical models are employed to capture pixel relationships within an image collection \cite{hu2021multi,jiang2020co,jiang2019multiple,jiang2019unified,wei2019deep,zhang2020adaptive}, followed by the extraction of co-salient objects characterized by consistent features. Some approaches leverage supplementary object saliency details to identify salient objects prior to implementing CoSOD \cite{jin2020icnet,zhang2021summarize,zhang2020coadnet}. Other methodologies focus on delineating shared attributes among input images \cite{fan2021group,zhang2021deepacg,zhang2020gradient,su2023unified,le2017co,zhu2023co,fan2021re,li2023discriminative,ge2022tcnet,zheng2023gconet+,wu2023co,xu2023co}, complementing semantic information with classification data. Comprehensive insights into CoSOD can be found in related surveys \cite{cong2018review,fan2020taking,zhang2018review}.

\textbf{Unsupervised segmentation:} Multiple approaches in unsupervised semantic segmentation leverage self-supervised feature learning methods \cite{ji2019invariant,li2021contrastive,van2020scan,cho2021picie,wang2022self}. Other works tackle unsupervised co-segmentation \cite{jerripothula2016image,li2018unsupervised,hsu2018co,amir2021deep,chakraborty2015site} and CoSOD \cite{zhang2016detection,hsu2018unsupervised}, where Li et al. \cite{li2018unsupervised} rank image complexities using saliency maps for unsupervised co-segmentation. Hsu et al. \cite{hsu2018co} propose an unsupervised co-attention model, and in \cite{hsu2018unsupervised}, their unsupervised graphical model jointly handles single-image saliency and object co-occurrence in CoSOD. Recently, Liu et al. \cite{liu2023self} introduced a self-supervised CoSOD model using an unsupervised graph clustering algorithm for detection, refining sample affinity with pseudo-labels. Additionally, Xiao et al. \cite{xiao2023zero} presented a zero-shot CoSOD approach that is based on group prompt generation and
subsequent co-saliency map generation. Chakraborty et al. \cite{Chakraborty_2024_WACV} proposed unsupervised and semi-supervised CoSOD models using segmentation frequency statistics that leveraged pre-trained models to generate pseudo-labels for training. Although ZS-CSD and US-CoSOD improved unsupervised CoSOD performance, relying on several off-the-shelf components made them computationally heavy. Our method outperforms all of  these unsupervised models while maintaining a lightweight design with minimal computational parameters. 

The existing unsupervised CoSOD methods suffer from limited performance due to their reliance on handcrafted features and insufficient utilization of feature correspondences at multiple scales. Our study addresses this gap by introducing a self-supervised CoSOD approach that effectively harnesses feature correspondences at different scales to significantly enhance CoSOD performance.

\vspace{-0.2mm}

 \begin{figure*}[h]
\centering
\includegraphics[width = 12.2cm]{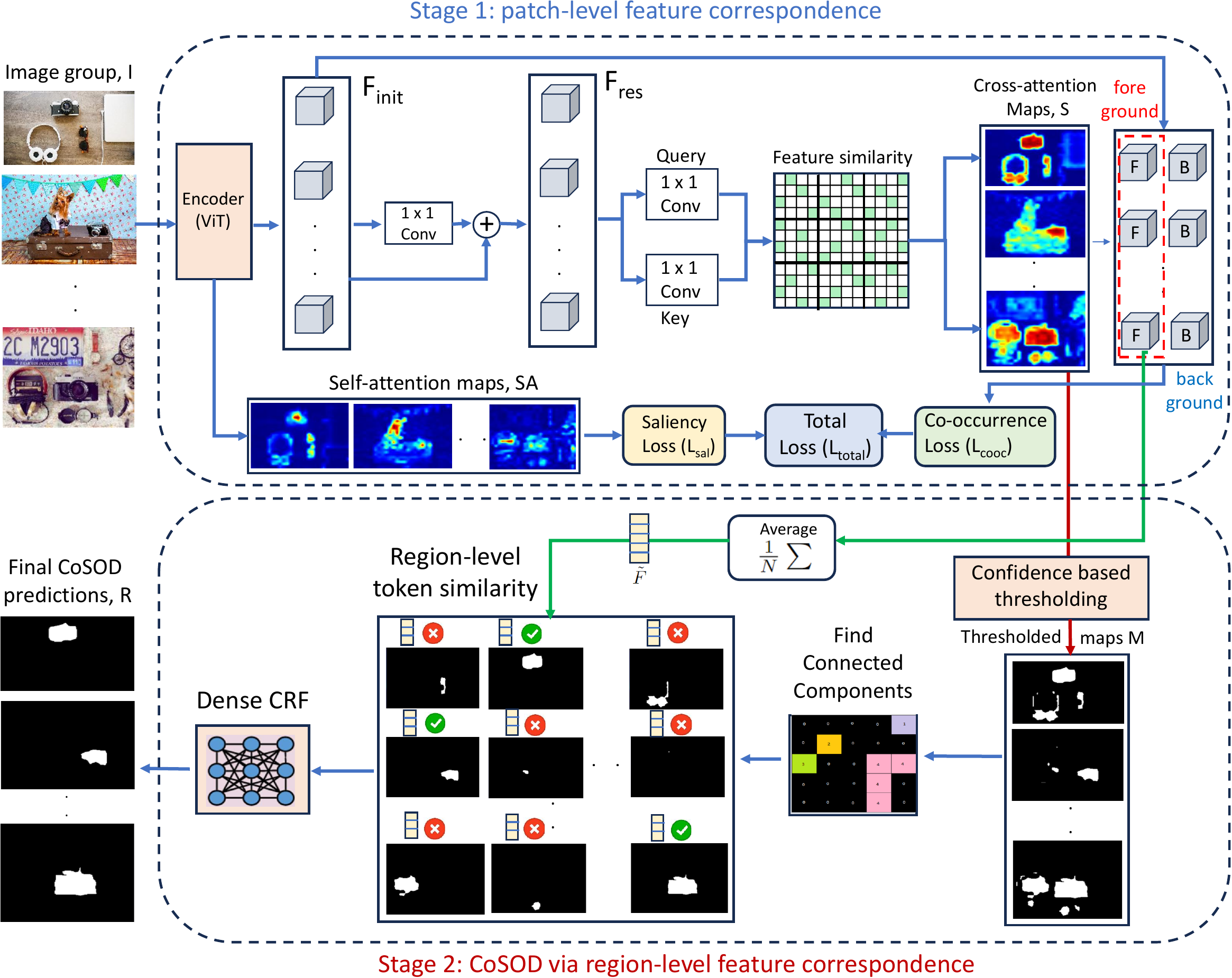}
\caption {The proposed two-stage self-supervised CoSOD model, SCoSPARC. In the first stage, we train a network that leverages the local ViT feature correspondences across all patches in the images in the group to obtain cross-attention heatmaps, which we further threshold using a confidence-based adaptive threshold to obtain an intermediate binary segmentation map. In the next stage, we refine these segmentations via region-level feature correspondence using the average foreground token obtained from the previous stage, followed by dense CRF-based segmentation refinement.}
\label{fig:unsup_figure}
\vspace{-8.3mm}
\end{figure*}
\section{Methodology}

Given a group of $N$ images $I = \{I_1, I_2, ..., I_N\}$ containing co-occurring salient objects of a specific class, CoSOD
aims to detect them concurrently and output their co-salient object segmentation masks. In self-supervised CoSOD, the goal is to predict the co-salient segmentations $\{\hat{y}_i\}_{i=1}^n$ without using any labeled data. 

Here, we describe our self-supervised CoSOD model, SCoSPARC  that employs ViT feature correspondences at both local and region levels to detect the co-salient objects in an image group.  

Fig.~\ref{fig:unsup_figure} depicts the pipeline of our SCoSPARC model. In the first stage, we leverage the patch-level (local) ViT feature correspondences across all patches in the images in the group to obtain the cross-attention map. We threshold this map using a confidence-based adaptive threshold to obtain an intermediate binary segmentation map. In the next stage, we refine these intermediate segmentations via region-level feature correspondence using the average foreground token obtained from the previous stage. Finally, we employ dense CRFs \cite{krahenbuhl2011efficient} to ensure spatial continuity in the predicted segmentations. We will describe each component in detail in the following subsections.

\subsection{Stage 1: Patch-level feature correspondences}
\label{sec:sec31}
Previous works on self-supervised learning (SSL) have shown that ViT~\cite{dosovitskiy2020image} models (pretrained on ImageNet) using methods such as 
DINO~\cite{caron2021emerging} can provide great features for segmentation tasks due to the explicit semantic information learned via SSL \cite{hamilton2022unsupervised,van2021unsupervised}. Motivated by this, we employ the pre-trained ViT trained using DINO as the feature encoder in our pipeline.

We first extract image patch features $\mathbf{x}^{pat}_n$ from an image $I_n$ in the image group using our ViT Encoder as: $\mathcal{F}_{init} = [\mathbf{x}_1^{\textit{pat}}, \dots, \mathbf{x}_N^{\textit{pat}}]$, where $\mathcal{F}_{init} \in \mathbb{R}^{N\times C \times H \times W}$ ($N$, $C$, $H$, $W$ are the number of images in the group, channel number, height, and width respectively) and $\mathbf{x}^{\textit{pat}}_{n} = Encoder({I_n})$. 

These features are processed by the residual block to generate residual features $\mathcal{F}_{res}$ as:
\begin{equation}
\centering
\mathcal{F}_{res} = \mathcal{F}_{init} + conv^{1\times1}(\mathcal{F}_{init}), 
\label{eq:residual}
\end{equation}
where $conv^{1\times1}$ represents for the $1\times1$ convolution layer and $\mathcal{F}_{res} \in \mathbb{R}^{N\times C \times H \times W}$. This layer when added to the DINO features generate strengthened residual features that better capture the complex relationships in the data. This makes training more efficient and allows faster network convergence.

First, we input the residual features $\mathcal{F}_{res}$ to our network. Next, we employ self-attention by utilizing two $1\times1$ convolution layers. These layers yield two distinct feature maps, namely the key map $K \in \mathbb{R}^{N\times C \times H \times W}$ and the query map $Q \in \mathbb{R}^{N\times C \times H \times W}$. After reshaping both $K$ and $Q$ to shape $ \mathbb{R}^{NHW\times C}$, we compute the feature similarity matrix $S$ as: 
\begin{equation}
\centering
S = \frac{1}{\sqrt{d}} K Q^\top,
\label{eq:seedsatt}
\end{equation}
where $S\in \mathbb{R}^{NHW\times NHW}$, $d$ = embedding dimension, and $\top$ denotes the transpose operation. Each row of $S$ represents feature token similarities between a patch (corresponding to the row) and all other patches of the $N$ input images. The feature similarity matrix S is then reshaped to shape $S \in \mathbb{R}^{N \times HW\times NHW}$. Next, we construct a 1D-map $S'_n \in \mathbb{R}^{HW}$ from the matrix $S$ for each image $I_n$ in the group by computing the row-wise mean of $S$ as:  $S'_n(p) = \frac{1}{NHW}\sum_{p'=1}^{NHW} S_{p'}$, where $p$ denotes a patch in $S_n$ and $p'$ is the corresponding index of patch $p$ (of $S'_n$) in the matrix S. Each 1D-map $S'_n \in \mathbb{R}^{HW}$ is next reshaped to a 2D-map $S_n \in \mathbb{R}^{H \times W}$. Maps $S_n$ are then separately normalized using min-max normalization. 

Although the pixel values of ground truth maps are either 0 or 1, those of the predicted feature similarity  maps $S$ contain intermediate intensity values (between 0 and 1), which indicate  uncertainty and noise in the predictions. To deal with these uncertain values, we employed a modified version of the Sigmoid function ~\cite{liao2022real} with a parameter $k$ that controls the steepness of  Sigmoid and encourages the map values of $S_n$ to be close to either 
0 or 1. $s_{th}$ is the confidence threshold. We represent the intermediate cross-attention maps $\mathcal{M}$ as:
\begin{equation}
  \mathcal{M}_n = \frac{1}{1 + e ^ {-k (S_n-s_{th})}},
  \label{eqn:DB} 
\end{equation}


Due to the nature of the task (co-salient object detection), we expect the detected co-salient regions across all images in the group to share similar feature representations in terms of object semantics and at the same time have a high saliency at an individual level. Therefore, we use a combination of two different loss terms to train our network in a self-supervised manner: (1) the co-occurrence loss $\mathcal{L}_{cooc}$ that measures the quality of the detected co-occurrent foreground regions between an image pair, and (2) the saliency loss $\mathcal{L}_{sal}$ that estimates the total saliency of the detected regions for an image. 
We define the co-occurrence loss $\mathcal{L}_{cooc}$ between images $I_n$ and $I_m$ as: 
\begin{equation}
\mathcal{L}_{cooc} = \sum_{n=1}^N \sum_{m=n}^N \frac{\exp({-d^{+}_{nm}})}{\exp({-d^{+}_{nm}}) + \exp({-d^{-}_{nm}})}
\end{equation}
\begin{equation}
d^{+}_{nm} = 1 - \cos({f(\mathcal{M}^f_n,\mathbf{x}^{pat}),f(\mathcal{M}^f_m,\mathbf{x}^{pat})})
\end{equation}
\begin{equation}
d^{-}_{nm} = 1 - (\cos({f(\mathcal{M}^f_n,\mathbf{x}^{pat}),f(\mathcal{M}^b_n,\mathbf{x}^{pat})}) + \cos({f(\mathcal{M}^f_m,\mathbf{x}^{pat}),f(\mathcal{M}^b_m,\mathbf{x}^{pat})}))
\end{equation}
where $f(m,\mathbf{x}^{pat}) = \frac{1}{n_p}\sum_{i=1}^{n_p} m_i \otimes \mathbf{x}^{pat}_i$ denotes the average ViT feature embedding corresponding to the mask $m$ and the patch descriptor  $\mathbf{x}^{pat}$ ($n_p$ = total number of patches in $m$). $\mathcal{M}_i^f = \mathcal{M}_i$ is the foreground mask (which is the cross-attention map) and $\mathcal{M}_i^b = 1-\mathcal{M}_i$ is the background mask corresponding to the image $I_i$. Here $cos$ denotes the cosine-similarity function.

Trained with the self-distillation loss \cite{hinton2015distilling}, the attention maps associated with the class token from the last layer of DINO  \cite{caron2021emerging} have been shown to highlight salient foreground regions \cite{caron2021emerging,wang2022tokencut,yin2022transfgu}. Their findings revealed that the attention heads of this model focus on significant foreground regions within an image. Motivated by this observation, we consider the averaged attention map (across all attention heads) from DINO as the foreground object segmentation. First, we average the self-attention maps from the $n_{h}$ DINO attention heads to obtain the averaged  self-attention map $SA_i$ for an image $I_i$ as: $SA_i = \frac{1}{n_{h}}\sum_{j=1}^{n_{h}} AM_i^j$, where $AM_i^j$ is the attention map from the DINO attention head $j$ for the image $I_i$. Map $SA_i$ is normalized by min-max normalization. Subsequently, the saliency loss $\mathcal{L}_{sal}$ is computed as:
\begin{equation}
\mathcal{L}_{sal} = 1-\frac{1}{N}\sum_{n=1}^N \mathcal{M}_n \otimes SA_n,
\label{eq:loss_combo}
\end{equation}

\noindent The network is self-supervised using a  combination of the co-occurrence and the saliency loss terms as:
\begin{equation}
\mathcal{L}_{total} = \mathcal{L}_{cooc} + \lambda_{sal} \mathcal{L}_{sal}
\end{equation}
where $\lambda_{sal}$ is the weight accounting for the saliency factor. Minimizing the saliency loss maximizes the average saliency of the detections. For co-occurring non-salient objects, the saliency loss is low and training/inference proceeds via the co-occurrence loss.

\paragraph{Confidence based adaptive thresholding:} While we enforce the map intensity values to be close to 0 or 1 (as explained in the previous paragraph), we require an additional thresholding step to obtain binary segmentation masks for each image. We observed that a more confident attention map requires a lower threshold and conversely, a less confident map requires a higher threshold in order to accurately predict a binary segmentation map of the co-salient regions. Consequently, the default threshold value of 0.5, used by the existing segmentation models (via the  \textit{argmax} operator) does not produce the best performance. Informed by this observation, we adaptively threshold the predicted cross-attention map $\mathcal{M}_n$ based on the confidence of the detected regions. To this end, we first compute the average confidence of the detected regions in the map $\mathcal{M}_n$ and then select the threshold $th$ as:

\begin{equation}
c_M = \frac{1}{n_{conf}} \sum_{p: p \geq \overline{\mathcal{M}}} \mathcal{M}_p
\end{equation}

\begin{equation}
b_M = 1 - c_M 
\end{equation}
\begin{equation}
th = th_{0} + \alpha_c (b_M-\overline{b_M}) 
\label{eq:eq11}
\end{equation}

\begin{equation}
G_n = 
\begin{cases}
    1 & \text{if } \mathcal{M}_n \geq th \\
    0 & \text{if } \mathcal{M}_n < th
\end{cases}
\end{equation}

\noindent where  $n_{conf}$ is the number of confident map pixels i.e. pixels with an intensity greater than the average intensity value $\overline{\mathcal{M}}$. $c_M$ denotes the average per-pixel  confidence  of the predicted map M for an image $I_n$ in the image group. $b_M$ is the inverse of the average confidence value of map M and $\overline{b_M}$ denotes the average value of $b_M$ over the training dataset. We set the adaptive threshold $th$ as $(b_M-\overline{b_M})$ offset by an initial threshold $th_0$. Thus, we threshold each map $\mathcal{M}_n$ to obtain the segmentation mask $G_n$ for each image ${I_n}$ in the image group I. See supplementary for more details.

\begin{algorithm}
\caption{Stage 2: Region-level mask refinement}
 \label{alg:algorithm1}
 \begin{algorithmic}[1]
 \renewcommand{\algorithmicrequire}{\textbf{Input: }}
 \renewcommand{\algorithmicensure}{\textbf{Output:}}
 \REQUIRE Image group $I = \{I_n\}_{i=1}^N$, intermediate segmentation mask $G = \{G_n\}_{i=1}^N$  
 \ENSURE Refined segmentation mask $R = \{R_n\}_{i=1}^N$  
\\

  \STATE Obtain the average token embedding $F_{G}$ corresponding to the masks $G$ across all images as:\\
  $F_{G} = \frac{1}{N} \sum_{i=1}^N \frac{1}{Ar(G_i)} \sum x_i^{patch} \otimes G_i$, \\$Ar(A) = $  area of region A.

  \FOR {$i = 1$ to $n$}
    \STATE Apply \textit{connected component labeling} on mask $G_i$ to generate $L_i$  masks for each component. 

    \STATE Map $R_i \gets$ all-zero map (same size as $G_i$)\;
    \FOR {$j = 1$ to $L_i$}
          \STATE Obtain the feature embedding $F_{G_{ij}}$ corresponding to the mask region $G_{ij}$ by averaging the patch embeddings as:
          $F_{G_{ij}} = \frac{1}{Ar(G_{ij})} \sum x_i^{patch} \otimes G_{ij}$.

          \STATE Compute the similarity between the token embeddings $F_{G}$ and $F_{G_{ij}}$ as: $d = \cos(F_{G},F_{G_{ij}})$, $\cos$ is the cosine distance.


          \textbf{if} $d_f \geq d_{f}^{th}$ then:\\
            \hspace{2mm} $R_i = R_i \cup G_{ij}$ \\
          \textbf{end if}
          

    \ENDFOR

  \ENDFOR
  
 \RETURN Refined masks, $R = \{R_n\}_{i=1}^N$ 
 \end{algorithmic} 
 \end{algorithm}

\subsection{Stage 2: Region-level feature correspondences} 
The regions highlighted in the binary segmentation maps $S$ obtained from stage 1 do not always belong to the co-salient object in the image group as shown in Sec.~\ref{sec:results}. This is because the patch-level feature correspondences fail to capture the region-level semantics of the co-occurring object. To  solve this problem, we eliminate regions whose feature are dissimilar with  the average consensus token representation i.e. the averaged token embeddings of the detected common foreground  regions. 

Our mask refinement algorithm is outlined in Algorithm \ref{alg:algorithm1}. First we obtain the average token embedding $F_{S}$ corresponding to the masks $S$ across all images by averaging the ViT patch embeddings. Next, we implement connected component labeling on the masks S in order to get sub-masks $L$ corresponding to the disconnected regions in these masks. For each sub-mask in each image, we compute the feature token similarity of the sub-mask with respect to the averaged token embedding $F_{S}$ and only retain sub-masks beyond a threshold similarity score $d_f^{th}$.  
\vspace{-2mm}
\paragraph{Postprocessing using denseCRFs:} Finally, we improve the co-salient segmentations $R_n$ obtained in the previous step by enforcing spatial
coherence and preserving object boundaries in the predictions using denseCRFs following previous
work \cite{hou2017deeply,li2016deep}  
for every image $I_n$ in the group.

\section{Experimental Results}
\label{sec:results}
\vspace{-1.0mm}

\subsection{Setup}
\noindent \textbf{Datasets and evaluation metrics:} For training our self-supervised SCoSPARC model, we used images from COCO9213 \cite{wang2019robust}, a subset of the COCO dataset \cite{lin2014microsoft} containing 9,213 images selected from 65 groups, and from the DUTS class dataset \cite{zhang2020gradient}
that contains 8,250
images in total distributed across 291 groups.  We evaluate our methods on three popular CoSOD benchmarks: CoCA \cite{zhang2020gradient}, Cosal2015 \cite{zhang2016detection}, and
CoSOD3k \cite{fan2020taking}. CoCA and CoSOD3k are 
challenging real-world co-saliency evaluation datasets, containing
multiple co-salient objects in some images, large appearance and scale variations, and complex backgrounds. 
Cosal2015 is a widely used  dataset for CoSOD evaluation. 

Our evaluation metrics include the  
Mean Absolute Error (MAE$\downarrow$) \cite{cheng2013efficient}, maximum F-measure ($F_{\beta}^{max}\uparrow$) \cite{achanta2009frequency}, maximum E-measure ($E_{\phi}^{max}\uparrow$) \cite{fan2018enhanced}, and S-measure ($S_{\alpha}\uparrow$) \cite{fan2017structure}.\\


\noindent \textbf{Implementation details:}
We use the ViT-B model (with patch size = 8 and patch descriptor dimension $d$ = 768) trained using DINO as our backbone feature extractor. For training, we set the sample size as the minimum of 24 or the total group size. At inference, all samples (resized to $224\times224$) in the group are input at once. We used the Adam optimizer to train our stage 1 network  for 80 epochs. The total training time is around 10 hours. The inference speed of the model is 20 FPS (without dense CRF) and 4 FPS (with dense CRF). All experiments are run on an NVIDIA Quadro RTX 8000 GPU. In Eq.~\ref{eqn:DB}, we empirically set the parameter $k$ to 6.66 and the threshold parameter $s_{th}$ to 0.65. In Eq.~\ref{eq:loss_combo} we set $\lambda_{sal}$ to 0.3. Increasing this value produced segmentations highlighting salient regions but not co-occurring. Decreasing this value highlighted commonly occurring background regions e.g. sky, roads, etc.  as being co-salient. We empirically set the embedding similarity threshold as $d_f^{th} = 0.75$ in Algorithm 1. In Eq.~\ref{eq:eq11} we empirically set $\alpha_c$ to 1 and $th_0$ to 0.5 (a widely used segmentation threshold). 
More details in the supplementary.

\begin{table*}[]
\begin{center}
\footnotesize
\renewcommand{\arraystretch}{1.0}
\setlength\tabcolsep{2.1pt}
\caption{Quantitative ablation studies of the proposed components in our model.}
\label{tab:ablation_modules}
\resizebox{12.5cm}{!}{%
\begin{tabular}{c|ccccc||cccc|cccc|cccc}
\hline
& \multicolumn{5}{c||}{Component}  & \multicolumn{4}{c|}{CoCA~\cite{zhang2020gradient}} & \multicolumn{4}{c|}{CoSal2015~\cite{zhang2016detection}} & \multicolumn{4}{c}{CoSOD3k~\cite{fan2020taking}} \\
ID &  \hspace{1.25mm} Co-oc. & Sal.  & CAT & RFC & D-CRF & $MAE \downarrow$  & $F_{\beta}^{max}\uparrow$ & $E_{\phi}^{max}\uparrow$ & $S_{\alpha}\uparrow$ & $MAE \downarrow$  & $F_{\beta}^{max}\uparrow$ & $E_{\phi}^{max}\uparrow$ & $S_{\alpha}\uparrow$ & $MAE \downarrow$  & $F_{\beta}^{max}\uparrow$ & $E_{\phi}^{max}\uparrow$ & $S_{\alpha}\uparrow$ \\
\hline
1 & \checkmark &  &  & &  & 0.105 & 0.565 & 0.756 & 0.678 & 0.075 & 0.851 & 0.892 & 0.823 & 0.077 & 0.801 & 0.868 & 0.793 \\
2 & \checkmark & \checkmark &  & &  & 0.105 & 0.564 & 0.754 & 0.678 & 0.072 & 0.853 & 0.895 & 0.830 & 0.075 & 0.810 & 0.869 & 0.798 \\
3 & \checkmark & \checkmark & \checkmark &  & &  0.105 & 0.567 & 0.756 & 0.679 & 
 0.069 & 0.840 & 0.893 & 0.832 & 
 0.069 & 0.802 & 0.878 & 0.808 \\
4 & \checkmark & \checkmark & \checkmark & \checkmark &  & 0.095 & 0.601 & 0.776 & 0.701 &  0.067 &  0.851 & 0.898 &  0.838 &  0.067 &  0.814 &  0.882 &  0.812 \\
\hline
5 & \checkmark & \checkmark & \checkmark & \checkmark & \checkmark &  \textbf{0.092} & \textbf{0.614} & \textbf{0.782} & \textbf{0.711} &
\textbf{0.062} & \textbf{0.869} & \textbf{0.905} & \textbf{0.851} &
\textbf{0.064} & \textbf{0.827} & \textbf{0.889} & \textbf{0.820} \\
\hline
\end{tabular}
}
\end{center}
\end{table*}

\begin{table*}[t]
\centering
\caption{
Comparison of our model with other state-of-the-art models on 3 benchmarks. We achieve state-of-the-art performance for  unsupervised CoSOD (upper block). Interestingly, our self-supervised SCoSPARC model outperforms recent supervised CoSOD methods (e.g. DCFM, CoRP, UFO, etc.) while being comparable to the SOTA (lower block).
}
\label{tab:sota_comp}
\setlength{\tabcolsep}{2pt}
\resizebox{12.5cm}{!}{%
\begin{tabular}{l|llll|llll|llll}
\toprule
 & \multicolumn{4}{c}{CoCA \cite{zhang2020gradient}} & \multicolumn{4}{c}{Cosal2015 \cite{zhang2016detection}} &
 \multicolumn{4}{c}{CoSOD3k \cite{fan2020taking}}\\

Method &
MAE$\downarrow$ & $F_{\beta}^{max}\uparrow$ & $E_{\phi}^{max}\uparrow$ & $S_{\alpha}\uparrow$ & 
MAE$\downarrow$ & $F_{\beta}^{max}\uparrow$ & $E_{\phi}^{max}\uparrow$ & $S_{\alpha}\uparrow$ &
MAE$\downarrow$ & $F_{\beta}^{max}\uparrow$ & $E_{\phi}^{max}\uparrow$ & $S_{\alpha}\uparrow$\\

\midrule

UCCDGO \cite{hsu2018unsupervised} (ECCV 2018) &
- & - & - & - & 
- & 0.758 & - & 0.751 & 
- & - & - & -\\

TokenCut \cite{wang2022self} (CVPR 2022) &
0.167 & 0.467 & 0.704 & 0.627 & 
0.139 & 0.805 & 0.857 & 0.793 & 
0.151 & 0.720 & 0.811 & 0.744\\
 
 DVFDVD \cite{amir2021deep} (ECCVW 2022) &
0.223 &  0.422 & 0.592 & 0.581 & 0.092 & 0.777 & 0.842 & 0.809 & 0.104 & 0.722 & 0.819 & 0.773\\

 SegSwap \cite{shen2022learning} (CVPRW 2022) &
0.165 &  0.422 & 0.666 & 0.567 & 0.178 & 0.618 & 0.720 & 0.632 & 0.177 & 0.560 & 0.705 & 0.608\\

SAM CSD 
\cite{liu2023self} (Elsevier CEE 2023) &
- & - & - & - &
0.092 & 0.782 & 0.847 & 0.782 &
0.108 & 0.703 & 0.810 & 0.723 \\

Zero-Shot CoSOD 
\cite{xiao2023zero} (ArXiV 2023) &
 0.115 & 0.549 & - & 0.667  &
 0.101 & 0.799 & - & 0.785 &
 0.117 & 0.691 & - & 0.723 \\

US-CoSOD
\cite{Chakraborty_2024_WACV} (WACV 2024) &
0.116 & 0.546 & 0.743 & 0.672 & 0.070 & 0.845 & 0.886 & 0.840 & 0.076 & 0.779 & 0.861 & 0.801\\

Group TokenCut &
 0.106 &  0.596 & 0.781 &  0.701 & 
 0.091 & 0.823 & 0.867 & 0.815 & 
 0.097 &  0.757 &  0.833 &  0.776\\

SCoSPARC (ours)  &
\textbf{0.092} & \textbf{0.614} & \textbf{0.782} & \textbf{0.711} &
\textbf{0.062} & \textbf{0.869} & \textbf{0.905} & \textbf{0.851} &
\textbf{0.064} & \textbf{0.827} & \textbf{0.889} & \textbf{0.823}\\

\midrule
\bottomrule
\\
\vspace{0.3mm}
GCAGC \cite{zhang2020adaptive} (CVPR 2020)  &
0.111 & 0.523 & 0.754 & 0.669 & 
0.085 & 0.813 & 0.866 & 0.817 &
0.100 & 0.740 & 0.816 & 0.785\\

GICD \cite{zhang2020gradient} (ECCV 2020) &
0.126 & 0.513 & 0.715 & 0.658 &
0.071 & 0.844 & 0.887 & 0.844 &
0.079 & 0.770 & 0.848 & 0.797\\

CoEGNet \cite{fan2021re} (TPAMI 2021)  &
0.106 & 0.493 & 0.717 & 0.612 &
0.077 & 0.832 & 0.882 & 0.836 &
0.092 & 0.736 & 0.825 & 0.762\\

GCoNet \cite{fan2021group} (CVPR 2021) &
0.105 & 0.544 & 0.760 & 0.673 &
0.068 & 0.847 & 0.887 & 0.845 &
0.071 & 0.777 & 0.860 & 0.802 \\

CSG \cite{zhang2022deep} (TMM  2022) &
0.106 & 0.532 & 0.739 & 0.671 &
0.062 & 0.841 & 0.895 & 0.845 &
0.087 & 0.753 & 0.842 & 0.788 \\

DCFM \cite{yu2022democracy} (CVPR 2022) &
0.085 & 0.598 & 0.783 & 0.710 &
0.067 & 0.856 & 0.892 & 0.838 &
0.067 & 0.805 & 0.874 & 0.810 \\

CoRP \cite{zhu2023co} (TPAMI 2023) &
- & 0.551 & - & 0.686 &
- & 0.885 & - & 0.875 &
- & 0.798 & - & 0.820 \\

UFO \cite{su2023unified} (TMM 2023) &
0.095 & 0.571 & 0.782 & 0.697 &
0.064 & 0.865 & 0.906 & 0.860 &
0.073 & 0.797 & 0.874 & 0.819 \\

MCCL \cite{zheng2023memory} (AAAI 2023)
 &
0.103 & 0.590 & 0.796 & 0.714 &
0.051 & 0.891 & 0.927 & 0.890 &
0.061 & \textbf{0.837} & 0.903 & \textbf{0.858} \\

GEM \cite{wu2023co} (CVPR 2023) &
0.095 & 0.599 & 0.808 & 0.726 &
0.053 & 0.882 & 0.933 & 0.885 &
\textbf{0.061} & 0.829 & \textbf{0.911} & 0.853 \\

DMT \cite{li2023discriminative} (CVPR 2023) &
0.108 & 0.619 & 0.800 & 0.725 &
\textbf{0.045} & \textbf{0.905} & \textbf{0.936} & \textbf{0.897} &
0.063 & 0.835 & 0.895 & 0.851 \\

GCoNet+ \cite{zheng2023gconet+} (TPAMI 2023) &
\textbf{0.081} & \textbf{0.637} & \textbf{0.814} & \textbf{0.738} &
0.056 & 0.891 & 0.924 & 0.881 &
0.062 & 0.834 & 0.901 & 0.843 \\

\bottomrule
\end{tabular}
}
\end{table*}

\begin{table*}[t]
\centering
\caption{
Comparison of our model with the SOTA supervised CoSOD model, GCoNet+ using different amounts of labeled data for training.
}
\label{tab:gconetp_comp}
\setlength{\tabcolsep}{2pt}
\resizebox{12.5cm}{!}{%
\begin{tabular}{l|l|llll|llll|llll}
\toprule
 & & \multicolumn{4}{c}{CoCA~\cite{zhang2020gradient}} & \multicolumn{4}{c}{Cosal2015 ~\cite{zhang2016detection}} &
 \multicolumn{4}{c}{CoSOD3k ~\cite{fan2020taking}}\\

Method & Label &
MAE$\downarrow$ & $F_{\beta}^{max}\uparrow$ & $E_{\phi}^{max}\uparrow$ & $S_{\alpha}\uparrow$ & 
MAE$\downarrow$ & $F_{\beta}^{max}\uparrow$ & $E_{\phi}^{max}\uparrow$ & $S_{\alpha}\uparrow$ &
MAE$\downarrow$ & $F_{\beta}^{max}\uparrow$ & $E_{\phi}^{max}\uparrow$ & $S_{\alpha}\uparrow$\\

\midrule

GCoNet+ \cite{zheng2023gconet+} & 50\% &
0.133 & 0.534 & 0.753 & 0.661 &
0.074 & 0.842 & 0.889 & 0.842 &
0.079 & 0.783 & 0.865 & 0.808 \\

GCoNet+ \cite{zheng2023gconet+} & 75\% &
0.113 & 0.547 & 0.759 & 0.682 &
0.066 & 0.863 & 0.902 & \textbf{0.860} &
0.071 & 0.804 & 0.876 & 0.823 \\

SCoSPARC (ours) & 0\% &
\textbf{0.092} & \textbf{0.614} & \textbf{0.782} & \textbf{0.711} &
\textbf{0.062} & \textbf{0.869} & \textbf{0.905} & 0.851 &
\textbf{0.064} & \textbf{0.827} & \textbf{0.889} & \textbf{0.823}\\

\bottomrule
\end{tabular}
}
\end{table*}

\subsection{Quantitative evaluation}
\paragraph{Ablation Studies:} In Tab.~\ref{tab:ablation_modules} we ablate the performance of our model using the different components, namely, the co-occurrence loss (Co-oc.), saliency loss (Sal.), confidence based adaptive thresholding (CAT), region-level  feature correspondence (RFC), and dense CRF (d-CRF). We observe that the saliency loss is useful for the Cosal2015 and the CoSOD3k test datasets. This could be attributed to the fact that CoCA focuses
more on segmenting the common objects in complex contexts, while Cosal2015 plays a more critical role in testing the ability of models to detect salient objects, as highlighted by \cite{zheng2023gconet+}. Nevertheless, we use the saliency loss in order to have a more generalized model. The confidence based adaptive thresholding (CAT) step in row 3 leads to an improved performance across  most metrics compared to using a fixed threshold of 0.5 in rows 1 and 2. Our region-level feature correspondence leads to a consistent improvement in performance by all metrics and across all the three test datasets. Finally, the dense CRF based post processing step leads to a consistent improvement in performance across all metrics and datasets. While dense CRFs improved segmentation performance of our model, even without this we outperform the SOTA unsupervised US-CoSOD  (Tab.~\ref{tab:sota_comp}) by a significant margin (see  Tab.~\ref{tab:ablation_modules}).

\begin{figure*}[t]
\centering
\includegraphics[width = 12.7cm]{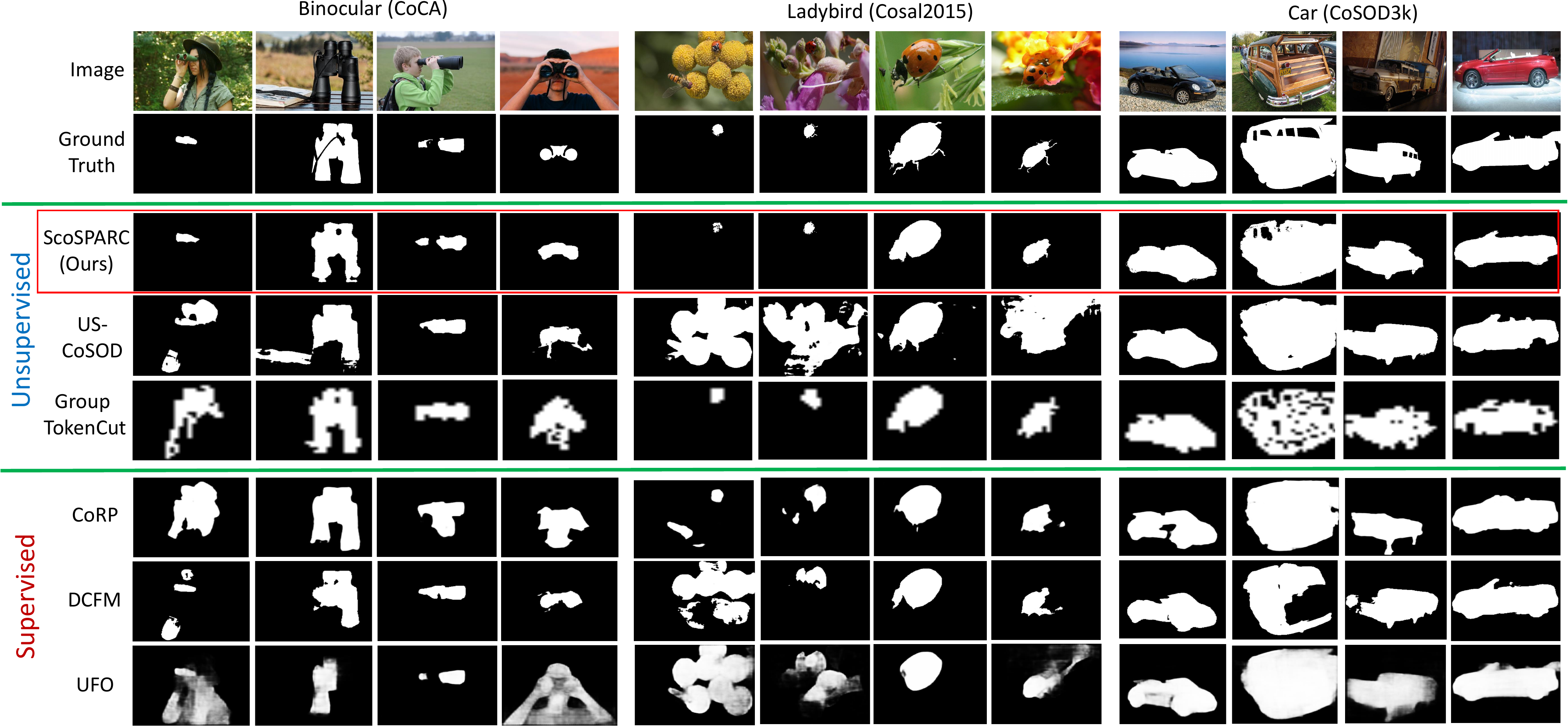}
\caption {Qualitative comparison of the performance of different baselines with our SCoSPARC on three image groups, each selected from the CoCA, Cosal2015, and CoSOD3k datasets. Our model produces the most accurate segmentations.}
\label{fig:qual_comp}
\vspace{-1.3mm}
\end{figure*}

\paragraph{Comparison with the state-of-the-art (SOTA) methods:} In Tab.~\ref{tab:sota_comp} we compare the performance of our model with the existing unsupervised CoSOD models (upper block) as well as supervised CoSOD models (lower block). 

In the upper block in Tab.~\ref{tab:sota_comp}, we see that our SCoSPARC outperforms all existing unsupervised CoSOD models. We introduce a baseline, \textit{Group TokenCut}, a modified version of the popular TokenCut \cite{wang2022self} model used for single-image foreground segmentation. In Group TokenCut, we calculate the second smallest eigenvector of a graph (indicating the likelihood of a  token belonging to
a foreground object) constructed across all patch-level tokens in the image group (from all images), differing from TokenCut's second smallest eigenvector computation based on a single image. We outperform the SOTA for unsupervised CoSOD i.e. the US-CoSOD model \cite{Chakraborty_2024_WACV}, by a significant margin (we achieve a 13.7\% gain in the F-measure metric over US-CoSOD on the CoCA dataset). 

Interestingly, in the lower block in Tab.~\ref{tab:sota_comp}, we see that SCoSPARC outperforms several SOTA fully supervised CoSOD models such as DCFM \cite{yu2022democracy}, CoRP \cite{zhu2023co}, UFO \cite{su2023unified}, etc. Also, we outperform the recent MCCL \cite{zheng2023memory} and GEM \cite{wu2023co} models on the CoCA dataset in terms of $F^{max}_{\beta}$-measure and MAE. 

In Tab.~\ref{tab:gconetp_comp} we quantitatively compare the performance of our SCoSPARC model with GCoNet+ \cite{zheng2023gconet+} when limited data is available for training. Specifically, we evaluated GCoNet+ using 50\% and 75\% training labels i.e. we randomly selected a fraction of images from each image group in the training dataset as the labeled set. We find  that GCoNet+ has worse performance compared to SCoSPARC using 50\% labels across all metrics, and in the majority of metrics using 75\% labels. We attribute the poor performance of GCoNet+ to the fact that this model overfits to the training data when limited data is available for training. Other supervised models such as CoRP \cite{zhu2023co}, DCFM \cite{yu2022democracy}, and UFO \cite{su2023unified} also perform poorly compared to our model due to the same reason. Our self-supervised model, on the other hand, better leverages the patch and region feature correspondences within the images without relying on labeled training data (thus avoiding 
overfitting), which improves prediction performance.

In Tab.~\ref{tab:compute} we compare the inference speeds of our SCoSPARC (with and without dense CRFs) with other unsupervised CoSOD baselines, namely SegSwap \cite{shen2022learning}, DVFDVD \cite{amir2021deep}, and Group TokenCut. SCoSPARC without the dense CRF step achieves the highest inference speed, in terms of the frames-per-second (FPS). Tab.~\ref{tab:ablation_modules} and Tab.~\ref{tab:sota_comp} show that our model outperforms all SOTA unsupervised CoSOD models and remains competitive with the recent supervised CoSOD models even without the dense-CRF post processing step.

\begin{table*}[t]
\centering
\caption{
Comparison of the inference speeds of SCoSPARC  (with and without dense CRFs) with other unsupervised CoSOD baselines.  
}
\label{tab:cost}
\setlength{\tabcolsep}{2pt}
\resizebox{7cm}{!}{%
\begin{tabular}{l|l}
\toprule
\centering
Method &
Inference Speed (FPS)\\

\midrule

SegSwap \cite{shen2022learning} (CVPRW 2022) &
0.50  \\

DVFDVD \cite{amir2021deep} (ECCVW 2022) &
0.23  \\

Group TokenCut  &
0.05 \\

SCoSPARC (ours)  &
4.1 \\

SCoSPARC w/o d-CRF (ours)  &
20.5 \\

\bottomrule
\end{tabular}
}
\label{tab:compute}
\vspace{-5mm}
\end{table*}

\subsection{Qualitative evaluation}

In Fig.~\ref{fig:qual_comp} we qualitatively compare the CoSOD predictions from our self-supervised SCoSPARC model with different baselines on three image groups, each from the CoCA, CoSOD3k, and Cosal2015 datasets. We compare our model with the unsupervised models US-CoSOD \cite{Chakraborty_2024_WACV} and \textit{Group TokenCut}, and with the supervised models CoRP \cite{wu2023co}, DCFM \cite{yu2022democracy}, and UFO \cite{su2023unified}. We observe that our SCoSPARC generates more accurate segmentations compared to other baselines. The unsupervised US-CoSOD \cite{Chakraborty_2024_WACV} and the \textit{Group TokenCut} models produce masks with several undesirable image regions (\eg the non-co-occurring foregrounds) that are responsible for their poor performances. While the supervised CoRP and the DCFM models generate reasonable segmentations in most cases, they fail to accurately detect the small objects for certain instances. For example, for the \textit{Ladybird} group from Cosal2015, in columns 1 and 2, most methods including CoRP and DCFM produce several undesirable image regions (in the flowers) while failing to accurately segment the small sized ladybird. Our model does not suffer from such drawbacks. UFO \cite{su2023unified} produces diffuse segmentation maps, often  leading to noisy predictions. 
In columns 2 and 3 of the \textit{car} group, we see failure cases where all models 
including ours erroneously detect background regions inter-leaving the windows as being salient. More results in supplementary.

In Fig.~\ref{fig:qual_2} we visualize the intermediate maps from our SCoSPARC model, namely the self-attention map, $SA$ from the DINO ViT backbone in column 2, the cross-attention map $S$ (from stage 1) in column 3, the thresholded segmentation map $G$ (following confidence based adaptive threshold) in column 4, the final segmentation $R$ (from stage 2) in column 5, and the ground truth in column 6, for the \textit{handbag} image group from the CoCA dataset. In column 4 in Fig.~\ref{fig:qual_2}, we highlight the regions eliminated using our region-based feature correspondence step (in stage 2) using dashed yellow boxes. We observe that this step  only retains the image regions that correspond well with the semantic information of the co-occurring object (handbag in this case) while eliminating undesirable image regions initially detected using local feature correspondence in stage 1. More results in the supplementary.

\begin{figure}[t]
\centering
\includegraphics[width = 11.9cm]{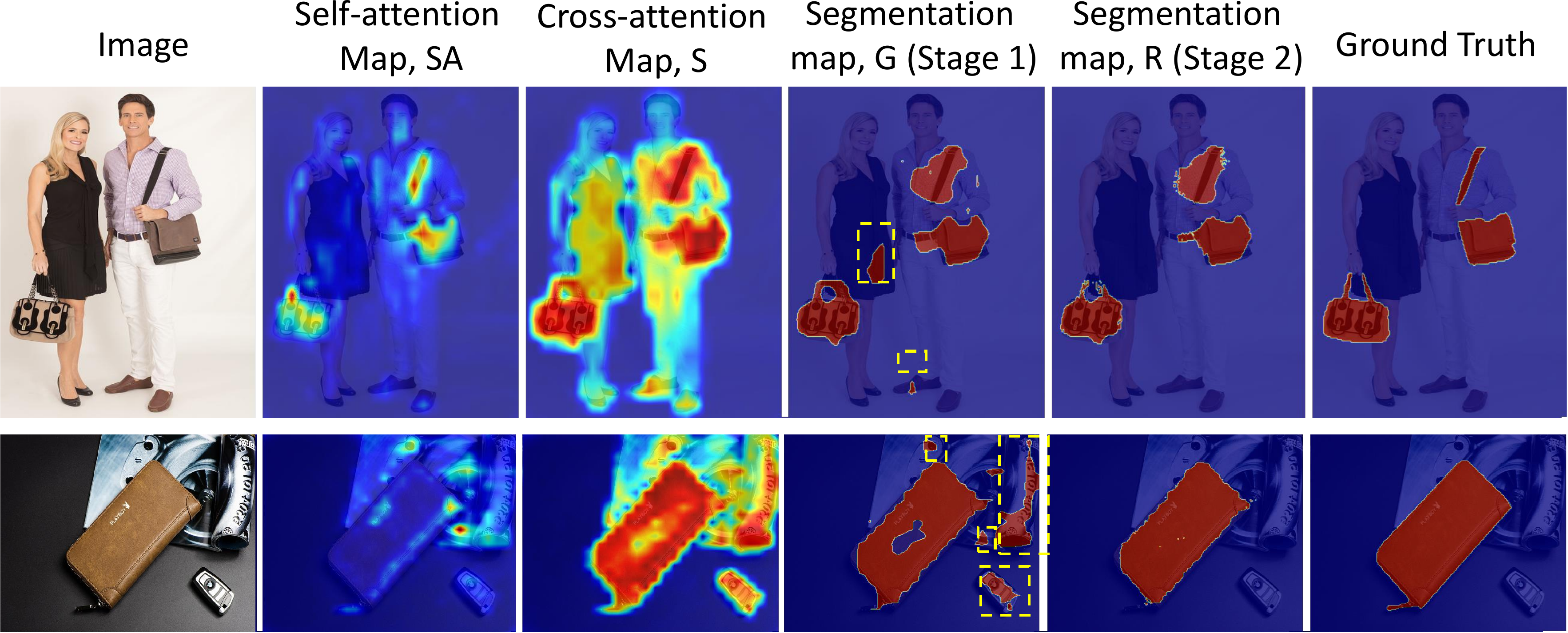}
\caption {Visualizations of  intermediate self-attention maps, cross-attention maps, and segmentation maps for two instances from the \textit{handbag} category from CoCA. The yellow boxes highlight the regions eliminated using stage 2 of our SCoSPARC model.}
\label{fig:qual_2}
\vspace{-2.3mm}
\end{figure}

\vspace{-0.1mm}
\section{Conclusion}
We presented a novel self-supervised approach for CoSOD based on mining feature correspondences at multiple scales within a group of images. Our model first finds local patch-level correspondences via a network trained to maximize co-occurrence and saliency of the detected regions in a self-supervised manner. We further employ a more global region-level correspondence to eliminate detected regions that do not align well with the consensus feature representation across the entire image group, which yields improved predictions. The proposed model outperforms all existing unsupervised methods and several popular supervised models for co-salient object detection. As future work, we would like to investigate the use of stable diffusion models (which has shown promising results for segmentation tasks) for self-supervised CoSOD using pseudo-labels from the proposed method. 
\clearpage
\setcounter{page}{1}

\noindent In this supplementary document, we provide details about our experiments and present more results from our study. The document is organized into the following sections:

\begin{enumerate}

    \item Section ~\ref{sec:quant_results2}: Additional quantitative results
    
    \item Section ~\ref{sec:qual_results2}: Additional qualitative results

   \item Section ~\ref{sec:imp_details2}: Additional implementation details
  
\end{enumerate}

\begin{figure*}
\centering
\includegraphics[width = 11.0cm]{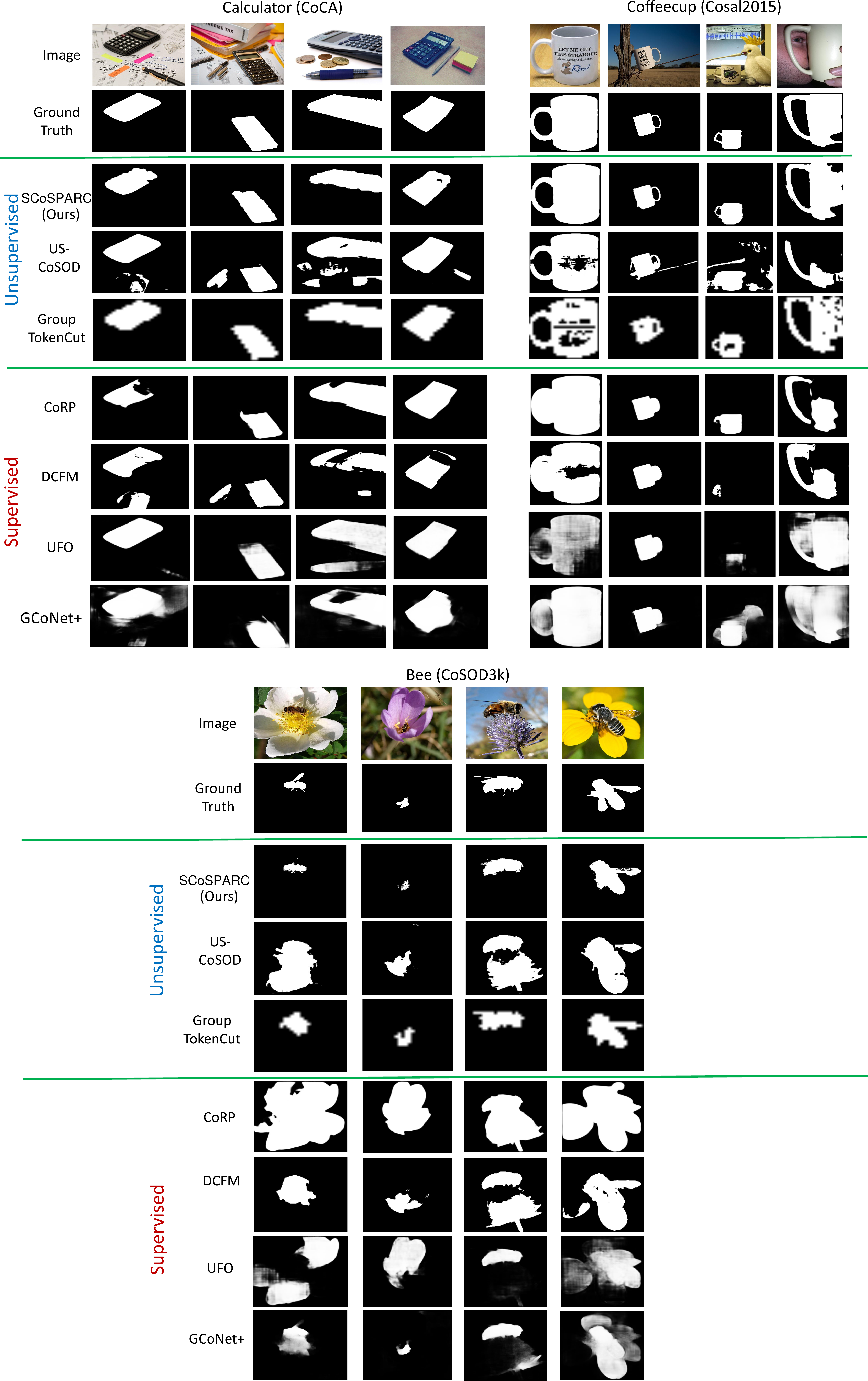}
\caption {Additional qualitative comparison of the performance of different baselines with our self-supervised CoSOD model on three image groups, each selected from the CoCA, Cosal2015, and CoSOD3k datasets. Our model produces the most accurate segmentations.}
\label{fig:qual_comp}
\vspace{-1.3mm}
\end{figure*}

\begin{figure*}
\centering
\includegraphics[width = 12.3cm]{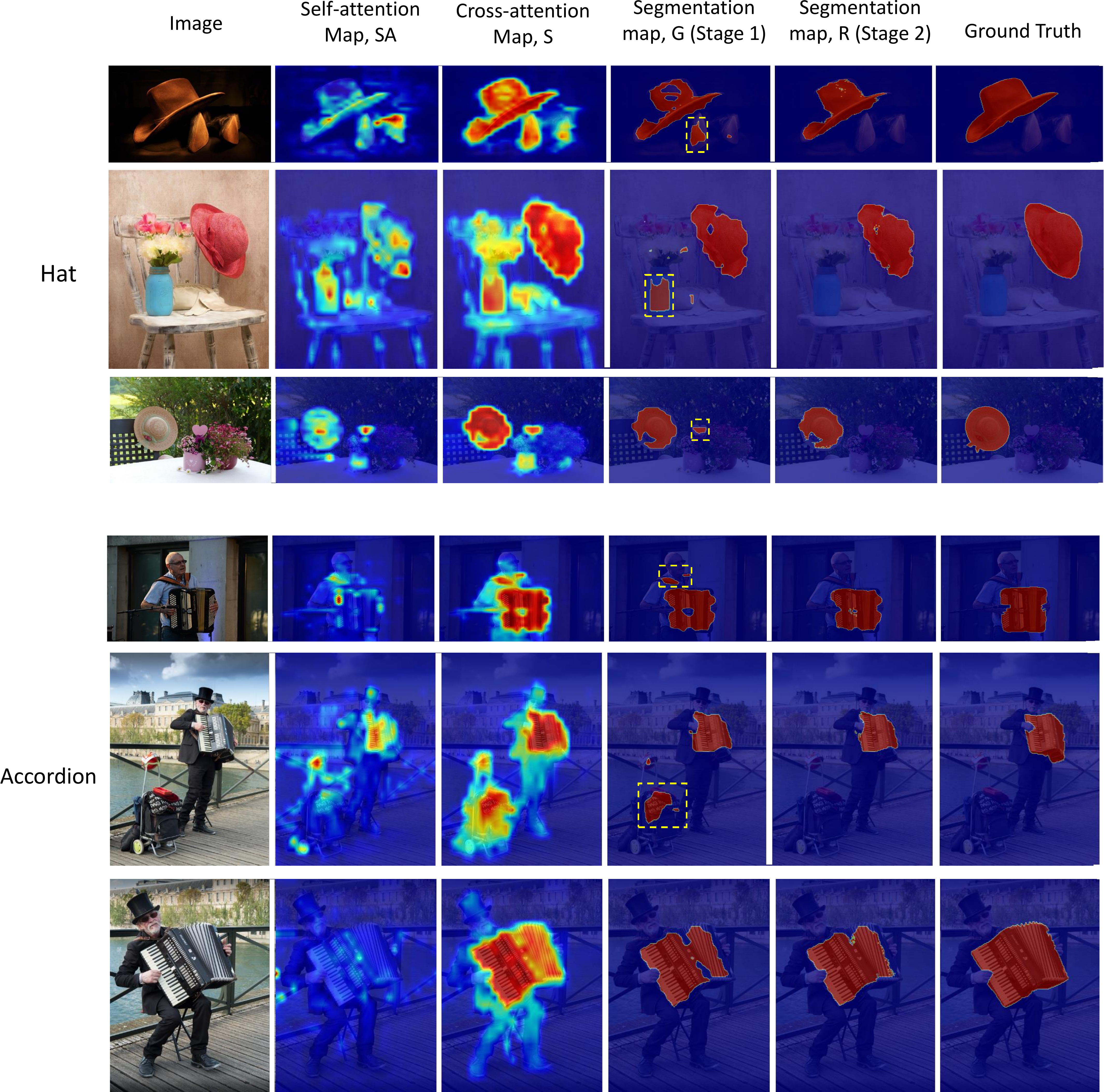}
\caption {Additional visualizations of the intermediate self-attention maps, cross-attention maps and segmentation maps for two instances from the \textit{handbag} category from the CoCA dataset. The yellow boxes highlight the regions eliminated using the stage 2 of our SCoSPARC model.}
\label{fig:int_att_maps}
\vspace{-1.3mm}
\end{figure*}

\begin{figure*}[t]
\centering
\includegraphics[width = 12cm]{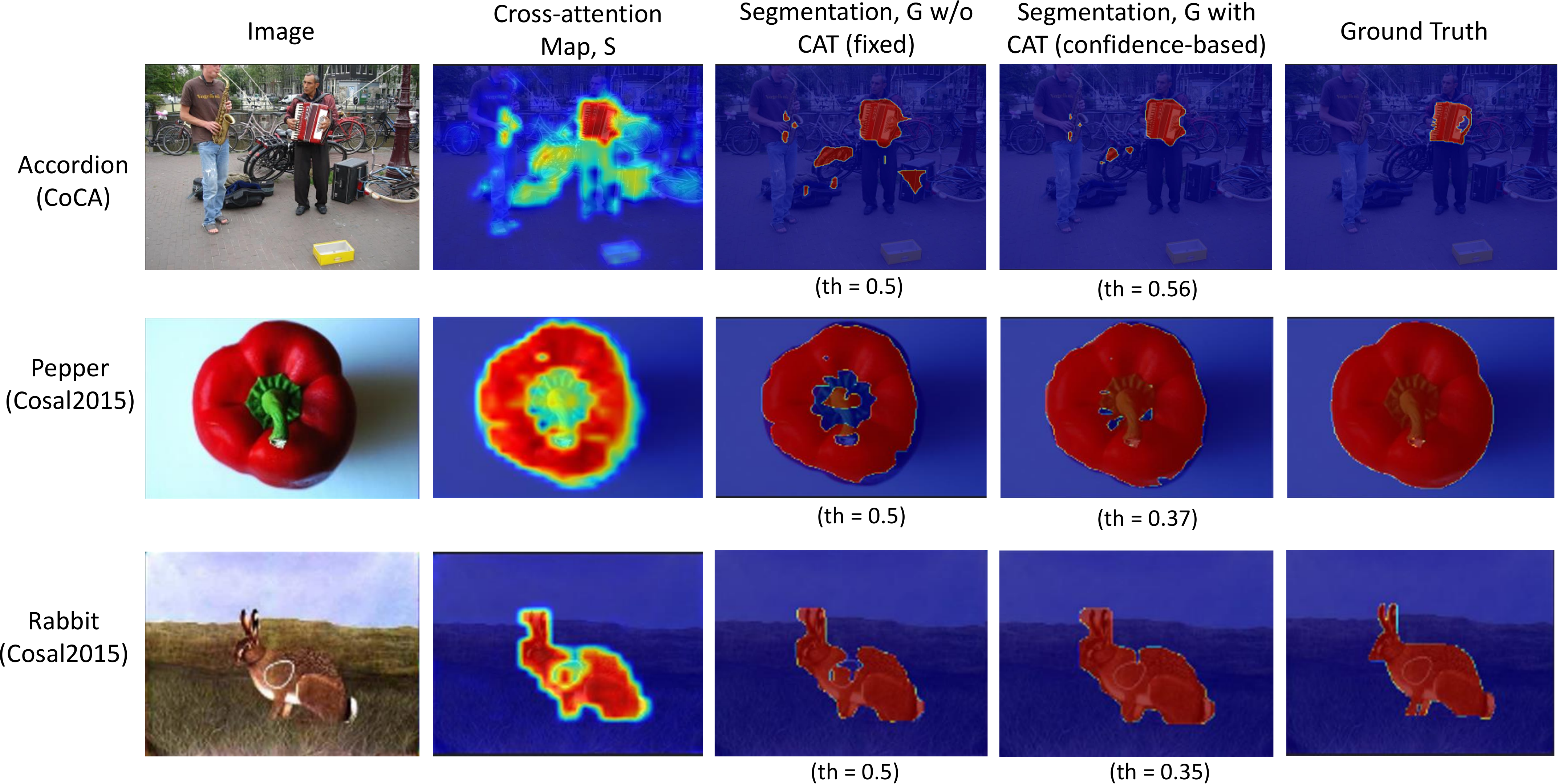}
\caption {Visualizations of the segmentation map, G (from stage 1) with and without our confidence-based adaptive thresholding (CAT) component. Our model with CAT produces more accurate segmentation predictions.}
\label{fig:int_maps}
\vspace{-1.3mm}
\end{figure*}

\begin{figure*}[t]
\centering
\includegraphics[width = 12cm]{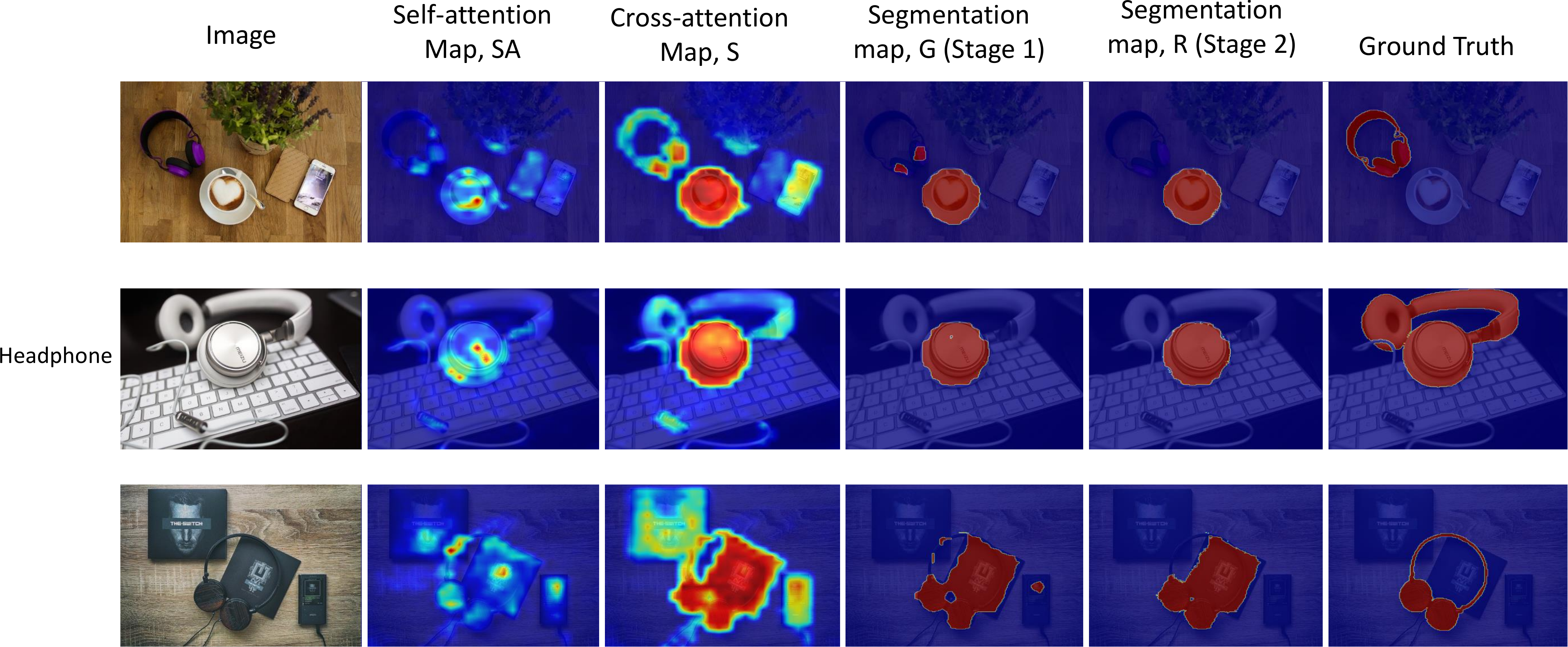}
\caption {Some failure cases of predictions on  the \textit{headphone} image group from CoCA.}
\label{fig:failure}
\vspace{-1.3mm}
\end{figure*}

\section{Additional quantitative results}
\label{sec:quant_results2}
\subsection{Performance using different training datasets}
In Tab.~\ref{tab:train_data} we compare the segmentation performance of our stage 1 self-supervised network when trained on the two datasets, namely COCO9213 \cite{wang2019robust} and DUTS-Class \cite{zhang2020gradient}. We observe that training on a combination of the two datasets  provides the best performance across all metrics and test datasets.

\begin{table*}
\centering
\caption{
Comparison of the segmentation performance of our stage 1 self-supervised network when trained on the two datasets, namely COCO9213 and DUTS Class. Training on a combination of the two datasets provides the best performance. 
}
\label{tab:train_data}
\setlength{\tabcolsep}{2pt}
\resizebox{12cm}{!}{%
\begin{tabular}{l|llll|llll|llll}
\toprule
 & \multicolumn{4}{c}{CoCA \cite{zhang2020gradient}} & \multicolumn{4}{c}{Cosal2015 \cite{zhang2016detection}} &
 \multicolumn{4}{c}{CoSOD3k \cite{fan2020taking}}\\

Method &
MAE$\downarrow$ & $F_{\beta}^{max}\uparrow$ & $E_{\phi}^{max}\uparrow$ & $S_{\alpha}\uparrow$ & 
MAE$\downarrow$ & $F_{\beta}^{max}\uparrow$ & $E_{\phi}^{max}\uparrow$ & $S_{\alpha}\uparrow$ &
MAE$\downarrow$ & $F_{\beta}^{max}\uparrow$ & $E_{\phi}^{max}\uparrow$ & $S_{\alpha}\uparrow$\\

\midrule

COCO9213 \cite{wang2019robust} &
0.115 & 0.532 & 0.737 & 0.659 & 
0.072 & 0.824 & 0.887 & 0.824 & 
0.074 & 0.781 & 0.866 & 0.800\\
\vspace{1.5mm}

DUTS-Class \cite{zhang2020gradient} &
0.105 & 0.555 & 0.760 & 0.674 & 
0.071 & 0.837 & 0.890 & 0.828 & 
0.070 & 0.801 & 0.878 & 0.805\\
\vspace{1.5mm}

COCO9213 \cite{wang2019robust} + DUTS-Class \cite{zhang2020gradient} &
0.104 & 0.567 & 0.756 & 0.679 & 
 0.069 & 0.844 & 0.894 & 0.832 & 
 0.069 & 0.806 & 0.878 & 0.808\\

\bottomrule
\end{tabular}
}
\end{table*}

\subsection{Performance using other encoder   backbones}
In Tab.~\ref{tab:encoder}, we show the effect of using different encoder backbones on the segmentation performance of our stage 1 self-supervised network. The ViT-Base encoder (embedding size = 768) with patch size = 8 provides the best performance, which we use in our final model. The convolution-based VGG-16 backbone \cite{simonyan2014very} has a significantly worse performance compared to the other ViT-based backbones. For ViT encoders, we observe that increasing the patch size from 8 to 16 (or reducing the prediction resolution) leads to a significant drop in performance \eg the F-measures on CoCA, Cosal2015 and CoSOD3k fall from 0.567, 0.844, 0.806 to 0.511, 0.760, and 0.664 respectively. Also, reducing the encoder's representation ability using a reduced embedding size (using  the ViT-Small backbone) while keeping the patch size same leads to a drop in performance \eg the F-measures on CoCA, Cosal2015 and CoSOD3k fall from 0.567, 0.844, 0.806 to 0.560, 0.830, and 0.752 respectively. 

\begin{table*}[t]
\centering
\caption{
The effect of using different encoder backbones on the segmentation performance of our stage 1 self-supervised network. The ViT-Base encoder with patch size = 8 provides the best performance.
}
\label{tab:encoder}
\setlength{\tabcolsep}{2pt}
\resizebox{12cm}{!}{%
\begin{tabular}{l|l|l|llll|llll|llll}
\toprule
 & \multicolumn{6}{c}{CoCA \cite{zhang2020gradient}} & \multicolumn{4}{c}{Cosal2015 \cite{zhang2016detection}} &
 \multicolumn{4}{c}{CoSOD3k \cite{fan2020taking}}\\

Encoder & Patch size & Embedding size &
MAE$\downarrow$ & $F_{\beta}^{max}\uparrow$ & $E_{\phi}^{max}\uparrow$ & $S_{\alpha}\uparrow$ & 
MAE$\downarrow$ & $F_{\beta}^{max}\uparrow$ & $E_{\phi}^{max}\uparrow$ & $S_{\alpha}\uparrow$ &
MAE$\downarrow$ & $F_{\beta}^{max}\uparrow$ & $E_{\phi}^{max}\uparrow$ & $S_{\alpha}\uparrow$\\

\midrule
VGG-16  & 16 & 512 &
 0.115 & 0.356 & 0.632 & 0.518 & 
0.205 & 0.475 & 0.553 & 0.505 & 
 0.173 & 0.468 &  0.572 &  0.517 \\

ViT-Base  & 16 & 768 &
0.116 & 0.511 & 0.743 & 0.640 & 
0.092 & 0.760 & 0.863 & 0.785 & 
0.119 & 0.664 & 0.792 & 0.724\\

ViT-Small & 8 & 384 &
0.105 & 0.559 & 0.755 & 0.667 & 
0.091 & 0.810 & 0.851 & 0.789 & 
0.081 & 0.779 & 0.852 & 0.778\\

ViT-Base  & 8 & 768 &
0.104 & 0.567 & 0.756 & 0.679 & 
 0.069 & 0.844 & 0.894 & 0.832 & 
 0.069 & 0.806 & 0.878 & 0.808\\

\bottomrule
\end{tabular}
}
\end{table*}












\section{Additional qualitative results}
\label{sec:qual_results2}
\vspace{-1.0mm}

\subsection{Comparison of CoSOD predictions}

In Fig.~\ref{fig:qual_comp} we qualitatively compare the CoSOD predictions from our SCoSPARC model with two unsupervised CoSOD models US-CoSOD \cite{Chakraborty_2024_WACV} and Group TokenCut and with four recent supervised models CoRP \cite{wu2023co}, DCFM \cite{yu2022democracy}, UFO \cite{su2023unified} and GCoNet+ \cite{zheng2023gconet+}. 

For the \textit{Calculator} class, we observe that the US-CoSOD model produces undesirable image regions as CoSOD detections. Our Group TokenCut baseline produces reasonably good detections in this case, although there are edge artifacts. The supervised models such as CoRP and DCFM produce incomplete segmentations in several instances (columns 1, 3 and 4). The DCFM model also segments undesirable image regions \eg the paper in column 1 and the pen in column 2. The UFO model inaccurately segments the pen as being co-salient in column 3. Finally, the GCoNet+ model, although being SOTA in supervised CoSOD, produces noisy predictions for this image group.
Our model produces the best results in general.

For the \textit{Coffecup} class, in column 1 we observe that all models except SCoSPARC produce either incomplete segmentations (\eg not detecting the textual regions on the cup) or inaccurately segment undesirable regions (\eg CoRP segments the background region between the cup handle). Similarly, in column 2, most baseline models inaccurately segment the background region between the cup handle. Also, in the third column, we see that our model produces the best results while other baselines produce inaccurate segmentations. In column 4, Group TokenCut produces comparable results to our predictions, while other models produce noisy segmentations.

For the \textit{Bee} class, most models except Group TokenCut inaccurately detect flower parts as being co-salient. While Group TokenCut produces reasonable segmentations in this case, the predictions of our SCoSPARC are more refined.

\subsection{Visualizations of intermediate maps}
In Fig.~\ref{fig:int_att_maps} we show additional visualizations of the intermediate self-attention maps, cross-attention maps and segmentation maps for two image groups, \textit{Hat} and \textit{Accordion} (three instances each) from the CoCA dataset. The yellow boxes highlight the regions eliminated using the stage 2 of our SCoSPARC model. We observe that undesirable image regions (i.e. non-co-salient regions highlighted by the yellow boxes) are eliminated in stage 2 segmentation predictions, $\mathcal{R}$ from our model using our region-level feature correspondence step (RFC).

\subsection{Visualizations of confidence-based adaptive thresholding results}
In Fig.~\ref{fig:int_maps}, we visualize the segmentation maps, G (from stage 1) with and without our confidence-based adaptive thresholding (CAT) component. We see that our model with CAT produces more accurate segmentation predictions. In row 1 of Fig.~\ref{fig:int_maps} (for an instance from the \textit{Accordion} category), we see that the CAT step eliminates undesirable image regions using a higher threshold of 0.56 (determined via prediction confidence) compared to the segmentation obtained using a fixed threshold of 0.5 (widely used in segmentation tasks). On the other hand, for the categories \textit{Pepper} and \textit{Rabbit}, we see that lower threshold values of 0.37 and 0.35 produces better segmentations respectively, compared to the fixed 0.5 threshold. The different threshold values predicted by our CAT step are based on the different average confidence intensities of the confident regions in the cross-attention map, $\mathcal{S}$ for the three cases. For example, the per-pixel confidence value (within the confident regions) of the map S for the \textit{Accordion} category (row 1) is lesser than the per-pixel confidence values for the \textit{Pepper} and \textit{Rabbit} categories, which leads the algorithm to predict a higher threshold for \textit{Accordion} compared to the other two categories in rows 2 and 3. This results in improved segmentations.

\subsection{Failure cases}

In Fig.~\ref{fig:failure} we show some failure cases of our SCoSPARC model on the \textit{Headphone} image group from the CoCA dataset. In row 1 the model misses the headphone and instead highlights the cup and plate as the co-salient objects. In row 2 only one side of the headphone object has been accurately segmented while the model fails to detect the other half including the headband. We observe that a lower threshold on the cross-attention map, S could have produced an improved segmentation (highlighting all parts of the headphone), which our model fails to predict. In row 3, our model predicts certain  undesirable regions as being co-salient along with the headphone.

\section{Additional implementation details}
\label{sec:imp_details2}
\subsection{Training details}

We use the ViT-Base model (with patch size = 8 and patch descriptor dimension = 768) trained using DINO as our backbone feature extractor. We freeze the weights of this backbone for all of our experiments. See Tab.~\ref{tab:encoder} for more ablations on the encoder choice. For training, we set the sample size as the minimum of 24 or the total group size. We input images with size $224\times224$. Using the ViT-Base model with patch size = 8 produces co-attention maps with size ($\frac{224}{8},\frac{224}{8}$) = $28\times 28$.

We used the PyTorch deep learning library and  the Adam optimizer to train our stage 1 network. We set the learning rate to $10^{-4}$ and the weight decay parameter to $10^{-4}$. The total training time for SCoSPARC is around 10 hours for 80 epochs. All experiments are run on an NVIDIA Quadro RTX 8000 GPU.  

\subsection{Inference details}

At inference, all samples (resized to $224\times224$) in the group are input at once.  The inference speed of the model is 20 FPS (without dense CRF) and 4 FPS (with dense CRF).

For the dense CRF \cite{krahenbuhl2011efficient} post-processing step , we generated the unary operator directly  from the binary segmentation map, $\mathcal{R}$ from stage 2. We set the smoothness kernel parameter $\theta_{\gamma} = 10$ and the appearance kernel parameters, $\theta_{\alpha}$ and $\theta_{\beta}$ to $10$ and $3$ respectively.  

\bibliographystyle{splncs04}
\bibliography{egbib}


%
%
\bibliographystyle{splncs04}
\bibliography{egbib}
\end{document}